%% file: main.tex
\documentclass[11pt]{article}

\usepackage[preprint]{acl}
\ifdefined\XeTeXversion
  \usepackage{fontspec}
\else
  \usepackage{times}
  \usepackage[T1]{fontenc}
  \usepackage[utf8]{inputenc}
  \usepackage{inconsolata}
\fi
\usepackage{latexsym}
\usepackage{microtype}
\usepackage{graphicx}
\usepackage{booktabs}
\usepackage{amsmath}
\usepackage{amssymb}
\DeclareFontShape{OML}{cmm}{m}{it}{<5>cmmi5<6-8>cmmi7<8>cmmi8<9>cmmi9<10-12>cmmi10<12->cmmi12}{}
\usepackage{mathtools}
\usepackage{amsthm}
\usepackage[capitalize,noabbrev]{cleveref}
\usepackage{xcolor}
\usepackage{multirow}

\newcommand{\pp}{\,pp}
\newcommand{\figorbox}[2]{\includegraphics[width=#2]{#1}}
\newtheorem{proposition}{Proposition}
\title{Done, But Still Working: Anytime Evaluation of\\ When Coding Agents Should Stop}
\input{authors}
\date{}
\hypersetup{pdfauthor={Zerui Yang, Quan Xue, Miaomiao Dong, Chaorui Zhang, Xuanyu Zheng, Xiyuan Wang, Haolin Ruan, Xueyan Niu, Guangcheng Song, Weixi Zhang, Wei Han, Bo Bai},pdftitle={Done, But Still Working: Anytime Evaluation of  When Coding Agents Should Stop}}
\begin{document}
\maketitle
\input{sections/00_abstract}
\input{sections/01_intro}
\input{sections/02_method}
\input{sections/03_setup}
\input{sections/04_results}
\input{sections/05_related}
\input{sections/06_discussion}
\input{sections/07_limitations}
\input{sections/10_impact}

\bibliography{references}

\newpage
\appendix
\input{sections/appendix}

\end{document}

%% file: authors.tex
\author{
  \textbf{Zerui Yang}\thanks{Corresponding author}, \textbf{Quan Xue}, \textbf{Miaomiao Dong}, \textbf{Chaorui Zhang}\thanks{Corresponding author}, \\
  \textbf{Xuanyu Zheng}, \textbf{Xiyuan Wang}, \textbf{Haolin Ruan}, \textbf{Xueyan Niu}, \\
  \textbf{Guangcheng Song}, \textbf{Weixi Zhang}, \textbf{Wei Han}, \textbf{Bo Bai} \\
  {\normalfont Huawei Technologies Ltd.} \\
  {\normalfont\small Correspondence:
  \texttt{richard.yangpolyu@gmail.com},
  \texttt{zhang.chaorui@huawei.com}}
}

%% file: sections/00_abstract.tex
\begin{abstract}
Coding agents often continue working after their own tests pass. Does this additional work improve the patch, and how much could be saved by stopping earlier? We study these questions through anytime evaluation: replaying recorded runs, scoring intermediate patches with hidden tests, and measuring the cost incurred at each step. For policies that only stop a run or select an earlier patch, this allows paired comparisons without generating new trajectories, provided replay preserves the relevant states. Across about 6{,}000 runs of eight open models in two scaffolds on SWE-bench Verified, we find substantial differences in the value of continued work. The hindsight saving available at passing self-tests ranges from under 1\% to 57\%, depending on the model and scaffold. A rule that stops after the same patch passes twice saves 23\% in a preregistered replication, but the preregistered quality test does not establish non-inferiority. We explain these differences by measuring how often a rule fires, how often the patch changes afterwards, and whether those changes repair or break it. Fine-tuning agents to imitate the rule saves 30\% over six seeds, while quality preservation remains uncertain. Our results help identify both the opportunities and the risks of early stopping.
\end{abstract}

%% file: sections/01_intro.tex
\section{Introduction}
\label{sec:intro}

\begin{figure*}[t]
\centering
\includegraphics[width=\textwidth]{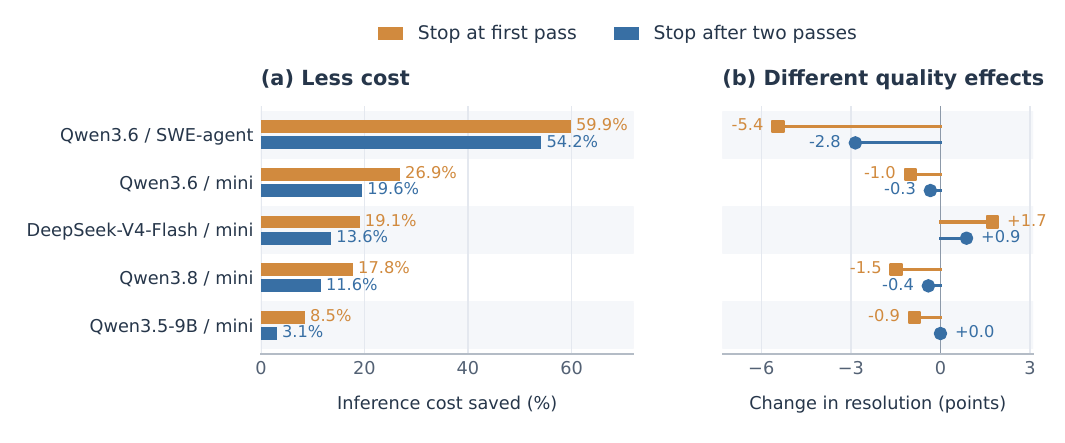}
\caption{Stopping saves cost, but its effect on patch quality depends on the agent. Two rules are compared with running to the end on the same recorded runs: stop at the first passing self-test, or after two passes on the same patch. Left: inference cost saved at $c=0.1$. Right: change in local resolution under corrected scoring $D'$ (percentage points). Point estimates use all runs of the five model--scaffold combinations shown in \cref{fig:bounds}, including derivation runs; these descriptive results are not the held-out confirmation test. ``mini denotes mini-SWE-agent.}
\label{fig:overview}
\end{figure*}

Coding agents search repositories, edit code and run tests to solve software issues \citep{jimenez2024swe,yang2024swe,wang2025openhands}. Testing helps an agent check its work, and the mini-SWE-agent prompt explicitly asks it to test edge cases before finishing \citep{yang2025minisweagent}. Yet a passing test does not necessarily end the run. The agent may run more tests, revise the patch or perform cleanup. These actions incur additional cost, and they may either fix remaining problems or introduce new ones. Figure~\ref{fig:overview} shows how two simple stopping rules trade cost against resolution across models and scaffolds.

This raises three questions. How much do agents spend after their tests pass? When does continued work improve the final patch? Can an agent stop earlier without losing those improvements? A benchmark score for the final patch cannot answer these questions because it does not show whether an earlier patch was already correct.

We address this gap with \emph{anytime evaluation} (\cref{sec:method}). We replay recorded commands in an executable sandbox and evaluate intermediate patches with hidden tests. We also calculate the cost incurred up to each command. The resulting trajectory shows when a patch becomes correct, whether it later breaks, and how much the remaining work costs. Hidden-test results from evaluation tasks are never shown to the agent. A separate training-only label ablation is identified explicitly.

This approach also lets us compare stopping rules on the same runs. If a rule only ends a run or selects an earlier patch, it does not change the actions that precede its decision. Provided those states are faithfully recorded or replayed, we can evaluate what the rule would have submitted without running the model again. Interventions that change the agent's behavior, such as fine-tuning, still require new runs.

We first measure the cost of continued work and the precision of paired comparisons. We then examine how much stopping could save, evaluate practical rules, and explain their quality differences. Finally, we test whether agents can learn to imitate a stopping rule.

\paragraph{Findings.} We study Qwen3.6-27B in mini-SWE-agent and SWE-agent with an executable sandbox on SWE-bench Verified tasks from Django and SymPy and from seven further repositories for which we built environments, and seven further open models, four from outside the Qwen family (\cref{sec:setup}). Primary tests and decision rules were specified before their corresponding runs or analyses; exploratory analyses are labeled separately (\cref{sec:results}).
\begin{itemize}
\item \textbf{Agents keep working after their own tests pass.} After its first passing self-test, Qwen3.6-27B runs 18 more commands, 42\% of the cost of those runs in mini-SWE-agent and 66\% in SWE-agent, mostly further tests, new test scripts and cleanup. In mini-SWE-agent it then submits the patch it already had in 80--87\% of these runs.
\item \textbf{Paired evaluation reduces uncertainty under explicit replay conditions.} Replay or snapshots match the recorded final tracked-file patch in 4{,}009 of 4{,}012 runs. This is an end-state check, not verification of every intermediate state. Comparing a rule with running to the end on the same runs has a standard error 2--8 times smaller than comparing independent runs, which is worth 4--60 times as many tasks. Detecting a 2-point change in resolution takes 56 tasks offline and about 1{,}200 online, more than SWE-bench Verified contains.
\item \textbf{Hindsight bounds put stopping rules in perspective.} Stopping at a passing self-test without losing quality can save at most 25\% of the cost in mini-SWE-agent and 57\% in SWE-agent. Stopping once the tests pass twice on one patch recovers 68--77\% of this bound for three models in mini-SWE-agent; in a preregistered replication on 386 new runs it saves 23\% at $-1.0$ points, but the preregistered test does not establish non-inferiority at our 2-point margin. Truncating anywhere could save 73--90\%, about half by stopping runs as soon as their patch is correct and half by abandoning runs that never succeed; both need information the agent does not have.
\item \textbf{A decomposition explains when stopping costs quality.} The change in resolution is the firing rate times the share of stopped runs whose patch changes later times the net repair per change. Only the last factor needs hidden tests; the cost side needs none. SWE-agent loses 2.8 points and mini-SWE-agent 0.3 because SWE-agent changes its patch after a pass in 40\% of stopped runs against 10\%, with a similar net repair per change. For DeepSeek-V4-Flash later changes break more than they repair, so stopping helps. Rules rarely fire for models that seldom see their tests pass. Plugging in the net repair of the other models predicts a held-out model's loss better than three baselines, but not significantly, and only under corrected scoring; we treat this prediction as exploratory.
\item \textbf{Scoring conventions change conclusions.} Scoring the worktree diff without staged changes understated the resolution of some models by up to 7 points, turned the 1.3-point gain of submitting the last passing patch into 0.3, and reversed the predictive result above.
\item \textbf{Stopping can be distilled.} Fine-tuned on its own runs cut at the twice-passed point, the agent stops by itself on unseen tasks: over six training seeds it saves 30\% at $-1.2$ points. In the historical diagnostic, its stop decisions at the base agent's passes differ from the rule in at most 2.2\% of cases; plain fine-tuning saves nothing, and stop labels at random or merely well-timed positions make it give up.
\end{itemize}

\paragraph{Contributions.}
\begin{enumerate}
\item Anytime evaluation for coding agents: replay with per-step scoring and pricing, the condition under which truncation and selection policies are evaluated exactly, the final-patch integrity check, and the precision gain of paired evaluation with the tasks it saves.
\item Hindsight bounds on what stopping and selection can gain, and an exact decomposition of their effect on resolution into firing, later patch changes and net repair, of which only the last needs hidden tests.
\item A preregistered evaluation of stopping and selection rules on eight open models and two scaffolds, with executable environments for seven additional SWE-bench Verified repositories, a scoring audit that reverses earlier conclusions, and negative results for prompted stopping, asking the agent, hindsight labels, a learned trajectory policy, risk-controlled rule choice and terse prompting.
\item Stop distillation as a case study of an intervention that changes behaviour: the agent learns to stop from its own runs labelled by a rule (two models, two scaffolds, three kinds of labels, three controls), and the decomposition separates the teacher's loss from imitation error.
\end{enumerate}

%% file: sections/02_method.tex
\section{Stopping as a Decision on Observed Prefixes}
\label{sec:method}

We want to compare finishing a run with stopping at an earlier point. For example, after a test passes, a rule may submit the current patch immediately; another rule may wait for a second passing test. To evaluate either choice, we need the patch available at that point, its correctness, and the cost already incurred. This section defines these quantities and explains which comparisons can be made from recorded trajectories.

\paragraph{Runs, policies and utility.}
A run issues commands $k=1,\dots,T$. After command $k$ the agent has observed the prefix $h_k$: the task, its own replies and the outputs so far. Let $\pi_k$ be the working-tree patch after command $k$, and $Q(\pi)\in\{0,1\}$ whether the task's hidden tests pass on $\pi$. A \emph{stop-and-select policy} $(\tau,\sigma)$ ends the run after command $\tau\le T$ and submits $\pi_\sigma$, $\sigma\le\tau$. It is \emph{prefix-measurable} if whether to stop at $k$, and which patch to submit then, are functions of $h_k$ alone; $\tau$ is then a stopping time of the agent's observations. Running to the end is $\tau=\sigma=T$. For a value $\lambda$ of a resolved task in cost units, the utility of a policy is
\begin{equation}
U(\tau,\sigma) = \lambda\,Q(\pi_\sigma) - \mathrm{cost}_\tau(c,p).
\label{eq:utility}
\end{equation}
We report cost and resolution separately rather than choose a single deployment value of $\lambda$; the utility defines the illustrative trade-offs in \cref{app:hazard}. Policies that read evaluation hidden tests are not deployable prefix-measurable rules; they are used for hindsight bounds and explicitly labeled training-only ablations.

\paragraph{Pricing a prefix.}
For requests $t=1,\dots,T'$ of a run, let $P_t$ be the prompt tokens of request $t$, $H_t\le P_t$ those served from the prefix cache, and $O_t$ the output tokens. The cost of the requests up to command $k$ is
\begin{equation}
\mathrm{cost}_k(c, p) = \sum_{t\le t_k}(P_t-H_t) + c\sum_{t\le t_k}H_t + p\sum_{t\le t_k}O_t,
\label{eq:cost}
\end{equation}
with $t_k$ the request that issued command $k$ and prices relative to an uncached input token. We report $c\in\{0.1,1\}$ and $p=4$. We count as cached only the tokens that the run's own previous request could have supplied, $H_t=\min(P_t, \hat H_t, P_{t-1})$, where $\hat H_t$ is the server-reported hit count, limiting credit for cache reuse supplied by concurrent runs (\cref{app:cost}).

\paragraph{Scoring every prefix.}
For each recorded trajectory, we check out the task's base commit in the same sandbox and execute the recorded commands in order, without calling the model. We record the patch $\pi_k$ after every command. In newer runs we record $\pi_k$ online instead. We score each distinct non-empty patch with the task's hidden tests (FAIL\_TO\_PASS and PASS\_TO\_PASS), applied as in SWE-bench. %
As an integrity check, replay or online snapshots matched the recorded final tracked-file working-tree patch in 4{,}009 of 4{,}012 runs across 16 run sets. We exclude the three mismatches; no replay execution failed. Final-patch agreement does not by itself verify every intermediate patch or test outcome. Our primary reanalysis uses $D'$, a limited correction of the recorded tracked-file diff $D$: when $D$ is empty but the recorded full diff contains tracked-file changes, we use those changes and replay the affected prefixes relative to the base commit. Partially staged non-empty diffs remain uncorrected. Earlier official scores used $D$. Official reports have now returned for all 307 distinct patches in the exported correction subset; one of its 93 tasks fails the gold self-check (\cref{app:officialreturn}). This subset does not cover every corrected analysis. This is not the official mini-SWE-agent submission protocol, which grades the patch file the agent writes when it submits (\cref{app:metric}).

\begin{figure*}[t]
\centering
\figorbox{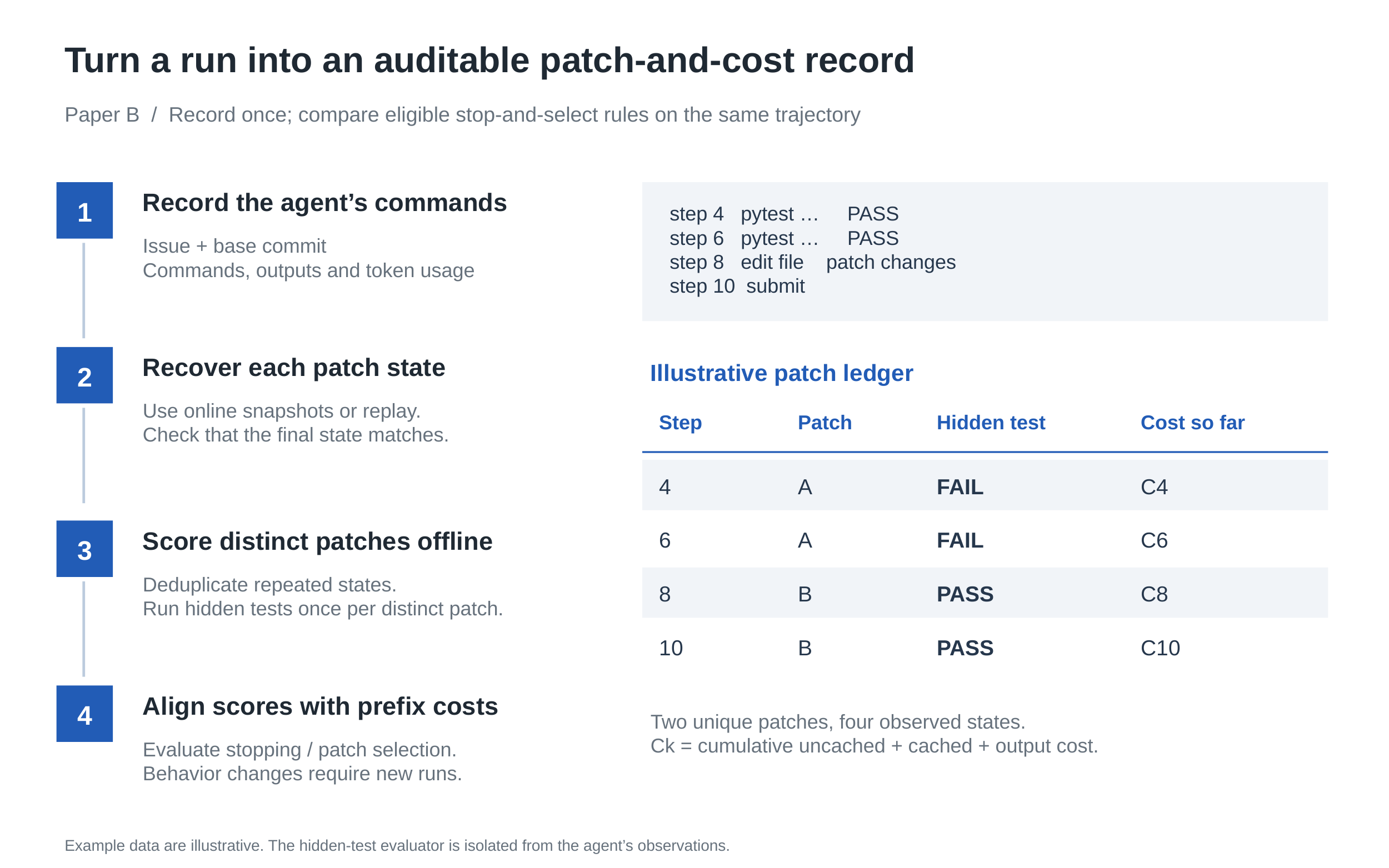}{\textwidth}
\caption{Anytime evaluation turns recorded commands into a patch-and-cost record. We recover intermediate patch states, score distinct patches with hidden tests offline, and align their scores with cumulative costs. The illustrative ledger contains four observations of two patches: repeated states reuse a score while cost continues to accumulate. Hidden-test results never enter the agent's input; interventions that change its behavior require new runs.}
\label{fig:pipeline}
\end{figure*}

Figure~\ref{fig:pipeline} shows the recording and scoring workflow.

\begin{proposition}[Condition for exact stop-and-select evaluation]
\label{prop:exact}
Assume the record preserves all observations, patch states and token counts used by the policy and evaluator, and the policy does not change the agent or environment before stopping. Let $(\tau,\sigma)$ be prefix-measurable and act only by ending the run and choosing among its patches. Then the submitted patch and the cost of a run with $(\tau,\sigma)$ applied online have the same joint distribution as $\pi_\sigma$ and $\mathrm{cost}_\tau$ computed on a run recorded without it.
\end{proposition}
Up to command $\tau$ the agent receives the same inputs whether or not the policy is present, so its commands, patches and token counts up to $\tau$ have the same distribution, and the policy's decision is a function of that prefix. Under these assumptions, such policies can be evaluated on the same recorded runs without generating new trajectories; population estimates still carry sampling error, and policy selection on evaluation runs can overfit. This is a truncation identity rather than a new general off-policy estimator; we call this \emph{anytime evaluation}. Because the policy and running to the end are compared on the same runs, the comparison is also far more precise than comparing independent runs (\cref{sec:precision}). The proposition does not cover anything that changes the prefix, such as a different prompt, a question to the agent or a fine-tuned model. Those we evaluate with new runs.

\paragraph{Rules.}
All rules see only what the agent sees: its commands, their outputs and the working tree. A command counts as a \emph{self-test} if it runs the repository's tests (\texttt{runtests.py}, \texttt{bin/test}, \texttt{pytest} or \texttt{unittest}) while the patch is non-empty. It \emph{passes} if its output reports passed tests and no failure (\cref{app:stop}). We study:
\begin{itemize}
\item \emph{First pass}: submit at the first passing self-test.
\item \emph{Twice passed}: submit at the first passing self-test whose patch equals that of the previous passing self-test, that is, after the agent has re-run its tests without changing the patch. Requiring $j$ passes on one patch gives the family \emph{$j$-th pass}.
\item \emph{Last pass} (selection sensitivity analysis): do not stop ($\tau=T$). Submit the patch of the last passing self-test if the agent changed the patch afterwards and never saw its tests pass again.
\item Baselines: step caps, stability rules (submit once a non-empty patch has not changed for $k$ commands) and a learned policy (\cref{app:stop}).
\end{itemize}
The hidden tests are used only for scoring and never enter the agent's input.

\begin{figure*}[t]
\centering
\figorbox{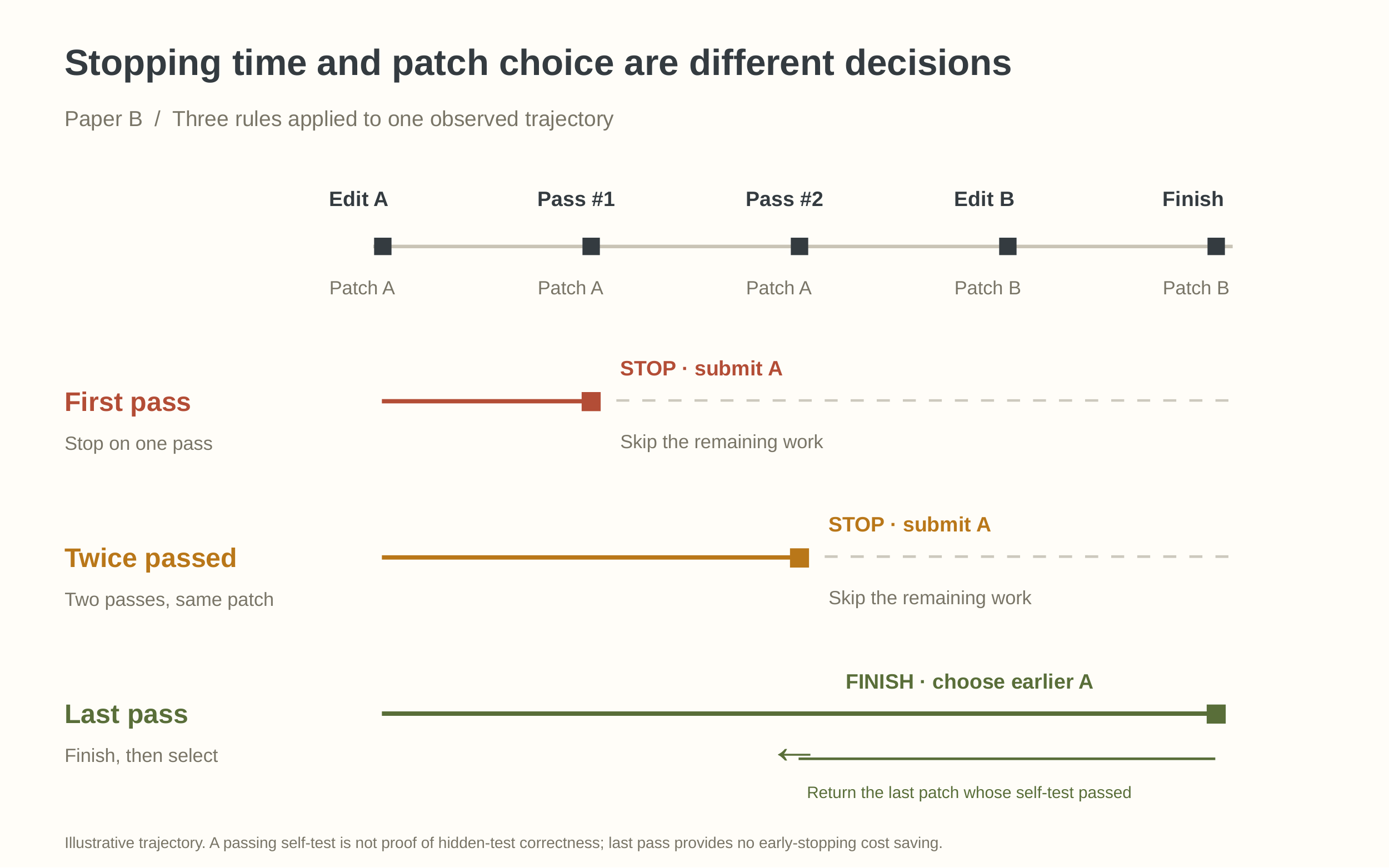}{\textwidth}
\caption{Three rules applied to one illustrative trajectory. First pass stops at the first passing self-test; twice passed waits for two passes on the same patch. Both submit patch A and skip later work. Last pass runs to completion, then chooses A because the later patch B has no passing self-test. This selection provides no early-stopping cost saving. Self-test success does not establish hidden-test correctness.}
\label{fig:rules}
\end{figure*}

Figure~\ref{fig:rules} contrasts stopping early with selecting an earlier patch after completion.

\paragraph{Hindsight bounds.}
Two policies that read the hidden tests bound what prefix-measurable ones can achieve. $O_{\mathrm{pass}}$ chooses, run by run, among the passing self-tests and the end: the earliest whose patch passes the hidden tests, otherwise the cheapest. $O_{\mathrm{any}}$ chooses in the same way among all commands and the end; a run whose patch never passes stops after its first command. Neither resolves less than running to the end. $O_{\mathrm{pass}}$ bounds every rule that stops or selects at passing self-tests, and the share of its saving that a rule attains is the rule's \emph{capture}.

\paragraph{Where stopping loses.}
For a rule that submits its current patch ($\sigma=\tau$), the change in resolution relative to running to the end is, run by run, $Q(\pi_\tau)-Q(\pi_T)$. It is zero if the rule does not fire, and also if the patch does not change after the stop. Averaging over runs gives the exact decomposition
\begin{equation}
\Delta Q = -\,r\cdot u\cdot\delta,
\label{eq:decomp}
\end{equation}
where $r$ is the share of runs in which the rule fires, $u$ the share of those whose final patch differs from the patch at the stop, and $\delta=E[Q(\pi_T)-Q(\pi_\tau)\mid\text{fired, changed}]$ the net repair per change: successes the rule would destroy minus those it would rescue. The saving $\mathrm{cost}_T-\mathrm{cost}_\tau$ is never negative. The decomposition is an algebraic identity on the scored records; interpreting it as an online policy effect requires the assumptions of \cref{prop:exact}. Only $\delta$ needs the hidden tests; $r$, $u$ and the saving can be read from a new agent's unlabelled runs. We use \cref{eq:decomp} to explain why rules behave differently across models and scaffolds (\cref{sec:decomp}).

\paragraph{Stop distillation.}
A rule decides only at passing self-tests, by a count. A model that reads the whole prefix could decide better, but labels from outcomes it cannot observe would bring hidden tests into training. We instead distill a prefix-measurable labelling rule $\ell$, such as twice passed, into the agent. On the agent's own runs of training tasks, $\ell$ fixes a stopping point $\tau_\ell$. A run gives a stop example, the prefix $h_{\tau_\ell}$ followed by a fixed submission $a_{\mathrm{stop}}$, and continue examples, prefixes $h_k$ at a few earlier commands $k\in K$ followed by the agent's own next reply $a_{k+1}$. We fine-tune on
\begin{equation}
\begin{split}
\mathcal{L}(\theta) = -\sum_{\text{runs}}\Big[&\log p_\theta(a_{\mathrm{stop}}\mid h_{\tau_\ell})\\
&+ \textstyle\sum_{k\in K}\log p_\theta(a_{k+1}\mid h_k)\Big].
\end{split}
\label{eq:distill}
\end{equation}
The first term teaches where to stop. The second is self-distillation, which keeps the agent's behavior elsewhere. The labels come only from the agent's own test output. The fine-tuned agent changes its own prefix, so we evaluate it with new runs. We diagnose its effect with two quantities: a \emph{stopping} component, measured by querying it at the passes in the base agent's recorded runs, and an online residual. The residual includes behavior changes and sampling variation; it is not an isolated causal estimate of behavior.

\paragraph{A diagnostic before training.}
The distilled agent imitates $\ell$, so its change in resolution should be close to the teacher's own $\Delta Q_\ell$ (\cref{eq:decomp}), measured on the base agent's training runs before any training. If the stopping part submits another patch than $\ell$ in a fraction $e$ of runs, it changes resolution by at least $\Delta Q_\ell-e$; the online residual has no such bound. We test this in \cref{sec:learned}.

%% file: sections/03_setup.tex
\section{Experimental Setup}
\label{sec:setup}

\paragraph{Agent and sandbox.}
We use mini-SWE-agent 2.4.6 \citep{yang2025minisweagent} with its official SWE-bench prompt, loop and submission protocol, and allow 250 steps. Commands run in an \emph{executable sandbox}: separate Linux namespaces, an unprivileged user, no network, per-command time limits, and a read-only Python environment for the task, so the agent can run the repository's tests but cannot install anything. Sampling uses temperature 0.7 and top-$p$ 0.8 with thinking disabled, and a seed fixed per task and repeat. Models are served with vLLM \citep{kwon2023efficient} with prefix caching.

\paragraph{Environments and tasks.}
We built per-version environments following the SWE-bench specifications, without Docker. For Django 3.0--5.0 and SymPy, the gold patch passes the hidden tests in 293 of 304 Verified tasks. For seven further repositories (pytest, Sphinx, xarray, pylint, requests, seaborn and Flask), we built 43 environments in which the gold patch passes on 78 of 106 tasks (\cref{app:envs}). We use only tasks whose gold patch passes, in four sets:
\begin{itemize}
\item \emph{Derivation}: 178 Django and SymPy tasks, two runs each with and without retrieved evidence (712 scorable runs). We designed all rules on these runs.
\item \emph{Fresh tasks}: 115 Django and SymPy tasks never used before, two runs each, plus online and prompt variants.
\item \emph{New repositories}: the 78 tasks above, two runs each.
\item \emph{New runs}: two further runs per task on the fresh tasks and new repositories (386 runs), used only for replication.
\end{itemize}

\paragraph{A second scaffold.}
To check that the behavior is not specific to mini-SWE-agent's prompt, we ran SWE-agent 1.1.0 \citep{yang2024swe} with its default configuration: function calling, its file viewer and string-replace editor, and its review step before submission. Its runtime server runs inside the same executable sandbox and talks to the host through a Unix socket, so every command and file operation of the agent stays in the sandbox. We changed only what the sandbox requires: the tools run on their own Python, commands time out after 60 seconds as in our other runs, pagers are disabled, and git accepts the repository despite its owner (\cref{app:envs}). Qwen3.6-27B ran each of the 193 fresh and new-repository tasks twice with the same sampling parameters (386 runs, preregistered). A proxy in front of the server records every request and tokenizes it exactly as sent, and we take a snapshot of the patch after every step. The rules, the test-command patterns and all thresholds are those frozen before.

\paragraph{Models.}
The main model is Qwen3.6-27B. We repeat the rules on the fresh tasks with Qwen3-Coder-30B-A3B, Qwen3.8-27B and Qwen3.5-9B, and with Gemma-4-26B-A4B, Gemma-4-31B-it and GLM-4.7-Flash from outside the Qwen family, two runs each. These three issue their commands as tool calls, one per turn, and their token counts are the prompt tokens returned by the server, because Gemma emits thinking segments that the tokenizer endpoint does not render (\cref{app:prereg}).

\paragraph{Quality and protocol.}
We additionally evaluate DeepSeek-V4-Flash in mini-SWE-agent on the same 115 fresh tasks, twice per task, using the text-action format with thinking disabled. All 230 runs completed; one run failed the recorded final-state consistency check, leaving 229 runs over 115 tasks for offline evaluation. This experiment tests the frozen stopping rules, not distillation. Together these are eight models from four families; Qwen3.6-27B is evaluated in both scaffolds, giving nine model--scaffold units in the cross-agent analysis.

Resolution is \emph{local}: the official hidden tests run in our sandbox and are parsed with the SWE-bench 4.1.0 log parsers. Historical official scoring used the same recorded working-tree patch $D$, not mini-SWE-agent's submitted patch file. Its agreement with local scoring therefore did not detect the shared omission of staged changes. We now report the limited correction $D'$ for the stopping table and distillation reanalysis (\cref{app:metric}); official scoring now confirms all 305 eligible distinct patches in the exported correction subset, and supplementary paired official estimates are reported in \cref{app:officialreturn}. This does not cover every intermediate patch or the later DeepSeek corrections. Unrecomputed historical diagnostics and their original official comparisons are explicitly distinguished from this reanalysis.

For stopping rules we required the upper bound of the 95\% CI of the cost change at $c=0.1$ to be at most $-10\%$, and the lower bound of the resolution change to be at least $-2$ points. Intervals use 2{,}000 task bootstrap resamples; runs of a task are averaged within a condition. The correction changes neither thresholds nor sampling design. Preregistration refers to dated internal research records written before the relevant run or analysis, not an independently timestamped public registry. Descriptive and exploratory analyses do not inherit confirmatory status from the experiment they analyze. %

%% file: sections/04_results.tex
\section{Results}
\label{sec:results}

\subsection{Agents keep working after their own tests pass}
In mini-SWE-agent, Qwen3.6-27B sees a passing self-test in 60\% of its runs. After the first one it runs 18 more commands, mostly further tests, new test scripts and cleanup, which account for 42\% of the cost of these runs and 27\% of all cost; in 80--87\% of them it then submits the patch it already had. In SWE-agent it passes a self-test in 90\% of runs and spends 66\% of their cost afterwards. DeepSeek-V4-Flash and Qwen3.8-27B spend about 40\% after a first pass in the 41\% and 46\% of runs where one occurs. Qwen3-Coder, Gemma-4 and GLM-4.7-Flash see one in at most 6\% of runs (\cref{app:models}).

\subsection{Paired evaluation improves precision}
\label{sec:precision}
The end-state check matched 4{,}009 of 4{,}012 final patches (\cref{sec:method}). This supports the integrity of patch reconstruction but does not prove all assumptions of \cref{prop:exact}; undetected intermediate-state differences remain a limitation. Because a rule and running to the end are compared on the same runs, most of the between-run variance cancels. \Cref{tab:precision} compares the standard error of a rule's effect with that of comparing two independent runs per task, one per arm, at the same budget of two runs per task. Offline evaluation is worth 4 to 60 times as many tasks; the gain is smallest in SWE-agent, where the rule changes more outcomes. On the 178 tasks of \cref{sec:learned}, detecting a 2-point change with 80\% power takes 56 tasks offline but 1{,}184 online, more than SWE-bench Verified contains, and still 296 with eight runs per task (\cref{app:power}). For this effect size and design, independent online comparisons need more tasks than the benchmark provides. Paired offline evaluation is therefore useful for the stopping part, while interventions that alter behavior still require new runs.

\begin{table}[t]
\caption{Standard error (points) of a change in resolution for the twice-passed rule against running to the end, two runs per task: offline on the same runs, or online from independent runs (rule on one run, end on the other). Factor: tasks saved, $(\mathrm{SE}_{\mathrm{online}}/\mathrm{SE}_{\mathrm{offline}})^2$. Qwen3.6 in mini-SWE-agent uses the 178 tasks of \cref{sec:learned}; rules that almost never fire are omitted.}
\label{tab:precision}
\centering
\footnotesize
\setlength{\tabcolsep}{4pt}
\begin{tabular}{@{}lrrrr@{}}
\toprule
Agent & Tasks & Offline & Online & Factor \\
\midrule
Qwen3.6, mini & 178 & 0.40 & 2.68 & 45.6 \\
Qwen3.6, SWE-agent & 193 & 1.28 & 2.58 & 4.0 \\
Qwen3.8 & 371 & 0.23 & 1.81 & 60.5 \\
Qwen3.5-9B & 115 & 0.62 & 4.15 & 45.1 \\
DeepSeek-V4-Flash & 114 & 0.88 & 3.94 & 20.2 \\
\bottomrule
\end{tabular}
\end{table}

\subsection{Hindsight bounds and stopping rules}
\label{sec:rules}
\begin{figure}[t]
\centering
\figorbox{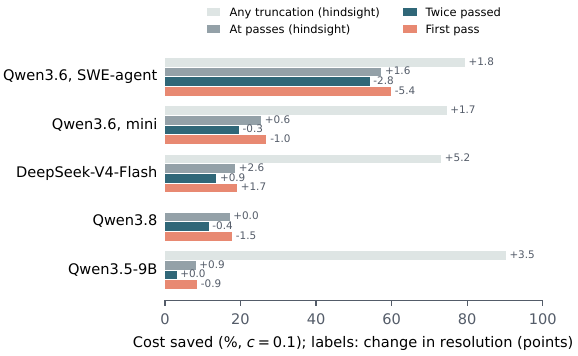}{\linewidth}
\caption{Cost saved by the hindsight bounds and by two rules, with the change in resolution (points) at the end of each bar; all runs of each agent under $D'$ (Qwen3.6, mini-SWE-agent: 1{,}484 runs including the derivation runs). $O_{\mathrm{any}}$ is omitted for Qwen3.8, whose intermediate patches are not all scored.}
\label{fig:bounds}
\end{figure}

\paragraph{What stopping can gain.}
Stopping or selecting at passing self-tests without losing quality ($O_{\mathrm{pass}}$) can save 57\% of the cost for Qwen3.6-27B in SWE-agent, 25\% in mini-SWE-agent, 17--19\% for Qwen3.8-27B and DeepSeek-V4-Flash, 8\% for Qwen3.5-9B and at most 5\% for the models that rarely see their tests pass (\cref{fig:bounds}). The twice-passed rule captures 68--77\% of this bound for Qwen3.6, Qwen3.8 and DeepSeek in mini-SWE-agent, and 95\% in SWE-agent, but there at a resolution 4.4 points below the bound. Truncating anywhere ($O_{\mathrm{any}}$) could save 73--90\%. For Qwen3.6 in mini-SWE-agent, 39 of its 75 points come from stopping successful runs as soon as their patch is correct, often before any test passes, and 36 from abandoning the 30\% of runs that never succeed. Both need information the agent does not have; the second suggests that giving up early is the larger open problem.

\begin{table}[t]
\caption{Paired rule comparisons on recorded runs of Qwen3.6-27B in mini-SWE-agent: change in local resolution (points) / change in cost at $c=0.1$, relative to running to the end, under $D'$. Rules were designed on the derivation runs; the fresh tasks (Django, SymPy) and the new repositories (78 tasks from seven repositories never run in our sandbox) test them, and the new runs (two further runs per task on both sets) replicate the preregistered tests. ``Last pass'' does not stop; it submits the last patch that passed a self-test if the agent changed it afterwards.}
\label{tab:stop}
\centering
\footnotesize
\setlength{\tabcolsep}{3pt}
\resizebox{\linewidth}{!}{%
\begin{tabular}{@{}lrrrr@{}}
\toprule
Rule & Derivation (712) & Fresh tasks (230) & New repos (156) & New runs (386) \\
\midrule
Self-test passed & $+0.4$ / $-23.9\%$ & $-3.9$ / $-31.1\%$ & $-1.3$ / $-27.3\%$ & $-1.8$ / $-30.5\%$ \\
Self-test passed twice on one patch & $+0.7$ / $-17.5\%$ & $-1.7$ / $-23.1\%$ & $-1.3$ / $-17.6\%$ & $-1.0$ / $-22.8\%$ \\
Last pass (no stopping) & $+0.4$ / $0\%$ & $+0.0$ / $0\%$ & $+0.0$ / $0\%$ & $+0.3$ / $0\%$ \\
\bottomrule
\end{tabular}}
\end{table}

\paragraph{Preregistered tests.}
\Cref{tab:stop} follows the twice-passed rule from design to replication. On new tasks and repositories it saves 18--23\% for $-1.7$ and $-1.3$ points. In the preregistered replication on 386 new runs it saves 22.8\% at $-1.0$ points. Although this point estimate is within the 2-point margin, the preregistered uncertainty-based test does not establish non-inferiority; an earlier analysis under $D$ had passed it (\cref{app:metric}). First-pass stopping saves 24--31\% but loses up to 3.9 points. Submitting the last passing patch gains at most 0.4 points under $D'$; the 1.3 points we first measured came largely from end patches scored as empty. Choosing $j$ for the $j$-th-pass rule by risk control with a 2\% loss budget selected the fourth pass, which saved too little (\cref{app:loss}).

\subsection{Why stopping loses: firing, change and repair}
\label{sec:decomp}
\begin{figure}[t]
\centering
\figorbox{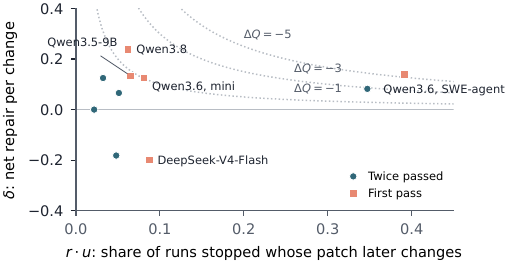}{\linewidth}
\caption{The decomposition $\Delta Q=-r\cdot u\cdot\delta$ (\cref{eq:decomp}) under $D'$. Horizontal: share of all runs in which the rule stopped and the patch later changed; vertical: net repair per change. Dotted lines: equal loss in resolution (points). Twice-passed points lie left of the first-pass point of the same agent.}
\label{fig:decomp}
\end{figure}

\paragraph{Scaffolds differ in how often they change, not in what changes buy.}
\Cref{fig:decomp} places each agent on the two factors of \cref{eq:decomp}. In SWE-agent the twice-passed rule loses 2.8 points and in mini-SWE-agent 0.3, yet the net repair per change is similar, $\delta=0.08$ and $0.07$ (0.14 and 0.12 for first pass). The difference is $u$: after the rule would have stopped, SWE-agent changes its patch in 40\% of runs and mini-SWE-agent in 10\%. SWE-agent runs nearly twice as many steps (93 against 51) and its tests 12--16 times per run instead of 4--9 (\cref{app:scaffold}). For DeepSeek-V4-Flash, $\delta$ is negative ($-0.18$ for twice passed, $-0.20$ for first pass): its later changes break more patches than they repair, so stopping raises resolution, by 0.9 and 1.7 points. Qwen3.8-27B changes its patch as rarely as Qwen3.6 but repairs more per change ($\delta=0.24$ at first pass), and its runs are short, so stopping saves less. For Qwen3-Coder, Gemma-4 and GLM-4.7-Flash, $r\le6\%$: the rules have nothing to act on, because these models seldom see the repository's tests pass.

\paragraph{Predicting the loss without labels (exploratory).}
Since $r$ and $u$ need no labels, the loss of a new agent can be predicted from its unlabelled runs and a $\delta$ pooled over other agents. Leaving out each of the nine agents in turn (preregistered), this prediction had a mean absolute error of 0.65 points for first pass and 0.31 for twice passed, against 1.05 and 0.50 for the best baseline, which scales the others' loss per firing by $r$ alone. The difference is not significant (intervals include zero), it fails for DeepSeek, whose $\delta$ has the opposite sign, and under $D$ the verdict reverses, because omitting staged end patches biases $\delta$ itself (\cref{app:metric}). The point estimates suggest a possible role for $u$, but neither a predictive advantage nor cross-model transfer is established.

\subsection{Telling, asking and teaching the agent to stop}
\label{sec:learned}
\paragraph{Telling or asking does not work.}
Replacing the official workflow's last step (test edge cases) with an instruction to submit once the relevant tests pass saved only 8.4\% (preregistered). Asked at its first passing self-test whether to submit, the agent said yes in 6--7\% of 439 cases, although in 61--72\% of the cases where it continued its patch already passed the hidden tests. A fresh judge of the same model saved 14.0\%, against 20.8\% for the twice-passed rule on the same runs (\cref{app:ask}).

\paragraph{Distilling the twice-passed rule.}
We applied stop distillation (\cref{eq:distill}) with twice passed as the labelling rule, fine-tuning Qwen3.6-27B with LoRA (rank 16, two epochs) on its own runs of the 115 fresh tasks: 215 stop examples and 896 continue examples (\cref{app:rl}). Run without any external rule on held-out tasks (preregistered), it cut cost by 27.9\% on the 78 new-repository tasks at $-0.6$ points and by 27.6\% on 178 further Django and SymPy tasks at $+2.5$ points. Over six training seeds on the 256 tasks it saved 30.0\% at $-1.2$ points ($-3.1$ to $+1.2$ by seed): the saving is robust, but the preregistered quality margin is not established. Official scoring gives the same qualitative conclusion on 254 fully paired tasks: cost changes by $-30.1\%$, while resolution changes by $-1.1$ points (\cref{app:officialreturn}). For Qwen3.8-27B the same recipe saved 43.2\% at $+0.2$ points.

\paragraph{Historical diagnostic of imitation and behavior ($D$).}
The following diagnostic retains the original $D$ scoring convention and has not been recomputed under $D'$. Letting each fine-tuned model decide at the passes of the base agent's recorded runs measures its stopping component under the fixed-prefix assumptions. These historical components must not be combined with the corrected effects above. In mini-SWE-agent it disagrees with the rule in at most 2.2\% of runs and stays within 0.6 points of it for every seed; the remaining seed variation appears in the online remainder, which also contains evaluation noise (\cref{tab:diag}). This does not identify a causal source of that variation. Plain fine-tuning on the same runs saved nothing (cost $+9.0\%$), stop labels at random earlier commands made the agent give up (2.2\% resolved), and labels at the rule's time in another run, off a pass, cost 14.6 points: the saving comes from stop labels at the evidence the rule uses. In SWE-agent the distilled agent saved 56.9\% and lost 3.2 points, the rule's own loss there; labels from the hidden tests could not be imitated from the prefix, and labels from exact advantages were worse teachers than the prefix-only rule (\cref{app:sweft,app:rl}).

%% file: sections/05_related.tex
\section{Related Work}
\label{sec:related}

\paragraph{Coding agents and their evaluation.}
SWE-bench \citep{jimenez2024swe} and its Verified subset \citep{openai2024verified} score the final patch of a run. Scaffolds range from purpose-built interfaces \citep{yang2024swe} and general platforms \citep{wang2025openhands} to bash-only agents \citep{yang2025minisweagent}. Benchmark differences are usually reported without error bars, and paired comparisons on the same questions are a standard remedy \citep{miller2024adding}. We keep each scaffold fixed, score every prefix of a run, and pair a policy with running to the end on the very same runs, so the two differ only where the policy changes a run's outcome.

\paragraph{Anytime algorithms and optimal stopping.}
Anytime algorithms return a usable answer whenever they are interrupted and are characterised by a performance profile of quality against time \citep{dean1988analysis,zilberstein1996using}; when to stop is then an optimal stopping problem \citep{chow1971optimal}. A coding agent is an anytime algorithm whose answer is its current patch. We measure its performance profile with hidden tests and price each prefix, and our hindsight bounds are those of a stopper that sees the whole run in advance.

\paragraph{Offline policy evaluation.}
Logged interactions can evaluate a new policy without deploying it: replay is unbiased when the logging policy is random \citep{li2011unbiased}, and importance weighting or doubly robust estimators correct for other logging policies \citep{dudik2011doubly}. Truncation and selection policies are a special case in which no correction is needed, because they never alter the prefix they act on; the remaining condition is that the record preserves the observations and states required by the policy and evaluator. Our final-patch match rate is a partial check of that condition.

\paragraph{Budget allocation in evolutionary search.}
EvalAlloc separates sparse feedback during scaffold evolution from targeted evaluation for final selection, reusing observations in paired successive halving \citep{luo2026evaluationbudget}. RelayEvolve allocates inexpensive-model search across trajectories and hands a curated population to a stronger model for refinement \citep{luo2026relay}. These methods allocate effort across candidate scaffolds or search populations. We study when to end a single coding-agent run and evaluate stopping rules on its recorded prefixes. Changing the scaffold or handing off to another model can change subsequent actions, so those interventions require new runs rather than our prefix-preserving offline comparison.

\paragraph{Stopping, overthinking and adaptive computation.}
Agents often keep acting after they have enough information. Overthinking in agentic tasks correlates with lower resolution \citep{cuadron2025danger}, and reasoning models spend many tokens after reaching an answer \citep{chen2024not}. Experience-driven early termination cuts the cost of software-engineering agents \citep{guo2026eetexperiencedrivenearlytermination}, turn caps at the 75th percentile of run lengths trade cost against solve rate \citep{gao2026more}, a learned monitor stops runs it predicts will fail \citep{liang2026stoploss}, and decision-theoretic control weighs verification against its cost \citep{papamarkou2026bayesian}. In NL2SQL, Memo-SQL combines structured decomposition with retrieval of successful queries and historical error--fix pairs to guide self-correction without training \citep{yang2026memosql}. Such work improves the correction process; our question is when further correction is worth its cost. Early-exit methods bound the quality lost by stopping early \citep{schuster2022confident}; choosing our rule with Learn then Test \citep{angelopoulos2025learn} under a 2\% loss budget was feasible but too conservative to save much (\cref{app:loss}). We study stopping after an observable self-test passes, which does not imply hidden-test success, diagnose the resulting quality changes, and use distillation as a case study. The empirical baseline comparison covers a fixed turn cap; we do not claim superiority to learned monitors or experience-driven termination methods that were not run here.

\paragraph{Patch selection.}
Sampling several runs and selecting a patch with generated tests or a learned verifier raises resolution \citep{xia2024agentless,ehrlich2025codemonkeys,pan2024training}. Our selection policies choose among the intermediate patches of a single run and cost nothing beyond the run itself.

\paragraph{Trajectory analysis.}
Trajectory studies locate decisive errors early and find that agents often fail to act on what they have found \citep{li2026codetracer,zhao2026failure,shu2026resolveratehidestrajectory}. CodeTracer's diagnostics include regressions in tests the agent itself ran \citep{li2026codetracer}. TallyMem combines source-preserving access to an agent's history with executable event aggregation to answer questions about past actions \citep{yang2026memorycompute}. Its target is historical question answering; ours is the cost and quality of stopping at an earlier patch. We instead score every intermediate patch with the hidden tests, which supports conditional evaluation of policies that only stop or select a patch, where the required intermediate scores are available.

\paragraph{Cost of agent runs.}
Input tokens dominate the cost of agentic coding and vary widely between runs of the same task \citep{bai2026ai}; in real use most of them are cache reads \citep{zhu2026tracelab}. Token reduction need not reduce cost under prefix caching \citep{weinberger2026token}, and context-management methods trade prompt length against cache hits \citep{lindenbauer2025complexity,xu2026tokenpilot}. Stopping removes whole turns, with their cached prefix and their output, and so saves under every price we tried.

%% file: sections/06_discussion.tex
\section{Discussion}
\label{sec:discussion}

\paragraph{Intermediate patches reveal what the final score misses.}
A final score tells us whether an agent eventually solved a task. It does not tell us when the solution appeared or how much the agent spent afterwards. Evaluating intermediate patches makes these questions measurable. It also allows paired comparisons of policies that only stop a run or select an earlier patch. These comparisons are conditional on faithful trajectories and do not replace online evaluation of interventions that change the agent's behavior.

\paragraph{A passing test is useful evidence, but not a guarantee.}
The value of stopping depends on what the agent would do next. We measure this through three quantities: how often the rule fires, how often the patch changes afterwards, and whether those changes improve correctness. The first two can be measured from ordinary agent runs. The third requires labeled evaluation, since later changes may help one model and harm another. This explains why the same stopping rule does not have the same quality effect across agents.

\paragraph{Learning to stop does not establish that stopping is safe.}
An agent can imitate a rule closely even when that rule sometimes stops too early. Before training, the rule should therefore be evaluated on recorded runs. After training, new runs are needed to measure any additional behavioral changes. Our six-seed results show reliable cost savings but do not rule out a meaningful quality loss. Finally, the patch used for scoring must be clearly defined. Omitting staged changes can make a non-empty patch appear empty and change conclusions about when an agent should stop.

%% file: sections/07_limitations.tex
\section{Limitations}
All eight models are open and locally served, four are from the Qwen family, and the runs use bounded budgets with thinking disabled where supported. We have not validated the results on frontier API agents or unrestricted reasoning settings. Tasks are restricted to repositories with executable environments and passing local gold checks; this selection limits generalization to the full benchmark. Test-command patterns can miss valid checks, especially custom scripts. Low firing rates therefore describe behavior under our detector, not an absence of all verification. A passing detected self-test is still a weak signal: hidden tests failed in about 30\% of the runs where it fired.

Exact stop-and-select evaluation is conditional on preserving the policy-relevant observations, intermediate patch states and cost accounting. Matching the final tracked-file patch does not prove that every replayed intermediate state or nondeterministic test outcome matches. For four models not every intermediate patch is scored, so $O_{\mathrm{any}}$ is not reported. The decomposition and bounds depend on the available local hidden-test scores. The cross-agent prediction is exploratory and sensitive to scoring conventions.

The primary correction $D'$ excludes new files and leaves partially staged non-empty diffs unchanged. It is not a complete base-relative patch protocol or an implementation of the official submitted-patch protocol. Official reports agree with local scores for the eligible exported correction subset, but do not cover all corrected analyses (\cref{app:officialreturn}). Historical diagnostics under $D$ are labeled separately and cannot be combined algebraically with $D'$ results. Full patch-definition robustness remains unresolved.

Rule replication reused the confirmation tasks. Distillation was trained for two models of one family; some held-out tasks come from the training repositories, and non-inferiority at the 2-point quality margin is not established. Historical imitation diagnostics do not guarantee the quality of a fine-tuned agent. Comparisons do not cover every published termination method. Cost means inference-token cost under the stated prices, excluding sandbox compute, monitoring overhead and wall-clock latency; no end-to-end latency saving is established here.

%% file: sections/10_impact.tex
\section*{Impact Statement}
This work studies when LLM coding agents should stop and which of their intermediate patches to submit. Its main practical consequence is lower inference cost. A stopping rule that relies on the agent's own tests passes less-verified patches to human reviewers where those tests are weak; we report how often the agent's passing tests disagreed with the hidden tests. All experiments use public benchmarks and open models; no human subjects or private code were involved.

%% file: sections/appendix.tex
\section{Preregistration and Deviations}
\label{app:prereg}
Each experiment was preregistered in a dated research log before it was run: the task set, rules or conditions, number of runs, primary tests, thresholds and decision rule. Results were written against these rules, including unsupported ones.

\begin{table*}[h]
\caption{Historical preregistered tests under the original $D$ convention. These original verdicts are preserved for provenance; the $D'$ reanalysis in \cref{app:metric} supersedes affected quality verdicts. Paired bootstrap over tasks (2{,}000 resamples); thresholds on 95\% CIs.}
\label{tab:prereg}
\centering
\footnotesize
\setlength{\tabcolsep}{4pt}
\resizebox{\textwidth}{!}{%
\begin{tabular}{@{}llll@{}}
\toprule
Experiment & Primary test & Expected & Verdict \\
\midrule
Stopping, 712 runs & learned policy: $\le-15\%$ cost, $\ge-0.01$ resolved, beats best simple rule & all three & not supported \\
Stopping, 115 fresh tasks & self-test rule: cost upper bound $\le-10\%$; resolved lower bound $\ge-0.02$ & both & cost supported; resolved not ($-0.030$) \\
Stopping, Qwen3-Coder, 115 fresh tasks & self-test passed twice: cost; resolved & as above & not supported (never fires) \\
Stopping, 78 tasks in 7 new repositories & self-test passed twice: cost upper bound $\le-10\%$; resolved lower bound $\ge-0.02$ & both & cost supported; resolved not ($+0.006$) \\
Stopping, 78 tasks in 7 new repositories & last passing patch, resolved & lower bound $>0$ & supported ($+0.032$) \\
Prompted stopping, 115 fresh tasks & cost upper bound $\le-10\%$; resolved lower bound $\ge-0.02$ & both & not supported ($-8.4\%$) \\
Asking at the first passing self-test, 3 task sets & cost upper bound $\le-10\%$; resolved lower bound $\ge-0.02$ & both & not supported (cost $-0.4\%$ to $+0.5\%$) \\
Stopping and last pass, Qwen3.8-27B, 115 tasks & twice-passed: cost, resolved; last pass: resolved & as above & not supported (0/3; $-13.6\%$, $-0.9$\pp, $0$) \\
Stopping and last pass, Qwen3.5-9B, 115 tasks & twice-passed: cost, resolved; last pass: resolved & as above & resolved supported, cost and last pass not ($-3.1\%$, $+0.4$\pp, $+0.9$\pp) \\
Replication, 386 new runs (Qwen3.6-27B) & last pass: resolved; twice-passed: cost, resolved & $>0$; $\le-10\%$; $\ge-2$\pp & supported (3/3) \\
Combined recipe online, 193 tasks & cost at $c=0.1$, $c=1$; resolved & $\le-15\%$, $\le-25\%$; $\ge-3$\pp & $c=1$ supported; $c=0.1$ and resolved not ($-3.5\%$; $-33.2\%$; $+0.1$\pp) \\
Terse prompting, 115 fresh tasks & cost upper bound $\le-5\%$; resolved lower bound $\ge-0.03$ & both & not supported ($-3.9\%$) \\
External judge, 193 tasks, 386 runs & stop at the judge's first yes: cost; resolved & $\le-10\%$; $\ge-2$\pp & cost supported; resolved not ($-14.0\%$ $-0.5$\pp) \\
Second scaffold (SWE-agent), 193 tasks, 386 runs & twice-passed: cost, resolved; last pass: resolved & $\le-10\%$; $\ge-2$\pp; $>0$ & cost supported; resolved and last pass not ($-54.2\%$; $-2.8$\pp $+0.3$\pp) \\
Stopping and last pass, Gemma-4-26B-A4B, 115 tasks (first attempt) & as for Qwen3.8-27B & as above & not run: smoke tests failed (format errors; token accounting) \\
Stopping and last pass, Gemma-4-26B-A4B, 115 tasks (second attempt) & as above & as above & not supported (fire in $\le$2\% of runs; $-0.1\%$, $0.0$\pp, $0.0$\pp) \\
Stopping and last pass, GLM-4.7-Flash, 115 tasks & as above & as above & not supported (fire in 3--6\% of runs; $-2.4\%$, $0.0$\pp, $0.0$\pp) \\
Fine-tuned stopping, 78 tasks in 7 new repositories, 156 runs & cost upper bound $\le-10\%$; resolved lower bound $\ge-0.02$ & both & cost supported; resolved not ($-27.9\%$ $+2.6$\pp) \\
Fine-tuned stopping, replication on 178 further tasks, 356 runs & as above & both & supported ($-27.6\%$ $+2.5$\pp) \\
Fine-tuned stopping in SWE-agent, 78 tasks in 7 new repositories, 156 runs & as above & both & cost supported; resolved not ($-56.9\%$ $-3.2$\pp) \\
Fine-tuned stopping, Qwen3.8-27B, 178 further tasks, 356 runs & as above & both & cost supported; resolved not ($-34.9\%$ $-0.3$\pp) \\
Fine-tuned stopping, Qwen3.8-27B, 78 new-repository tasks; pooled 256 tasks & as above; pooled resolved & both & cost supported; resolved not ($-56.0\%$; $+1.3$\pp pooled $+0.2$\pp) \\
Fine-tuned stopping in SWE-agent with hidden-test labels, 156 runs & as above & both & cost supported; resolved not ($-51.2\%$ $-4.5$\pp) \\
Fine-tuned stopping in SWE-agent with fourth-pass labels, 156 runs & as above & both & cost supported; resolved not ($-48.3\%$ $-2.6$\pp) \\
Loss-controlled choice of $k$ (Learn then Test, $\alpha=0.02$), mini and SWE-agent & held-out $E[L]$ bound $\le0.02$ and saving $\ge15\%$ / $10\%$ & both & not supported (3/3; $k=4$ chosen, saves 4\% / 4\% / 32\%) \\
Fine-tuned stopping with exact-advantage labels (agent's own tests), 178 tasks & vs.\ base: cost upper bound $\le-10\%$, resolved lower bound $\ge-0.02$; vs.\ twice-passed agent: cost upper bound $<0$, resolved lower bound $\ge-0.02$ & all four & 1/4 (cost vs.\ base $-30.8\%$; resolved $-0.6$\pp vs.\ agent $-4.3\%$, $-3.1$\pp) \\
Fine-tuned stopping with exact-advantage labels (hidden tests), 178 tasks & as above & all four & 2/4 (cost vs.\ base $-35.1\%$; resolved $-0.3$\pp vs.\ agent $-10.4\%$ $-2.8$\pp) \\
Exact-advantage labels on 78 new-repository tasks; pooled 256 tasks & cost; resolved (vs.\ base) & both & cost supported, resolved not (pooled: own tests $-28.6\%$, $+0.8$\pp hidden tests $-33.6\%$, $-1.2$\pp) \\
Fine-tuned stopping, second training seed, 178 tasks & cost upper bound $\le-10\%$; resolved lower bound $\ge-0.02$ & both & cost supported; resolved not ($-27.4\%$ $+0.3$\pp) \\
Fine-tuned stopping, training seeds 3--6, 178 tasks each & as above & both & cost supported 4/4 ($-25.5\%$ to $-32.6\%$); resolved 1/4 (seed 3: $+3.7$\pp others $-1.1$ to $+1.1$\pp) \\
Fine-tuned stopping, training seeds 2--6, 78 new-repository tasks; pooled 256 tasks & as above & both & cost supported 5/5 and 5/5; resolved 0/5 and 2/5 (pooled: seeds 3 and 6) \\
First fine-tuned model, new sampling seeds, 178 and 78 tasks; pooled 256 & as above & both & cost supported; resolved not (pooled $-30.0\%$; $-1.4$\pp $-3.9$\pp vs.\ its first evaluation) \\
First fine-tuned model vs.\ base model, both with new sampling seeds; pooled 256 & as above & both & cost supported; resolved not ($-34.9\%$; $-2.5$\pp) \\
Fine-tuned stopping, control without stop labels, 178 tasks, 356 runs & distilled minus control: cost upper bound $<0$; resolved lower bound $\ge-0.02$ & both & supported ($-33.6\%$ $+2.5$\pp) \\
Fine-tuned stopping, control with random stop positions, 178 tasks, 356 runs & distilled minus control: resolved & $>0$ & supported ($+63.5$\pp) \\
Fine-tuned stopping, control with stop positions matched in time but off a pass, 178 tasks, 356 runs & distilled minus control: resolved & $>0$ & supported ($+14.6$\pp) \\
SWE-agent, fourth-pass labels with continue examples at earlier passes: imitation gate & $e\le0.10$ and stopping part $\ge-2.5$\pp & both & not met ($e=0.127$; $-1.4$\pp); online run done as exploratory \\
\bottomrule
\end{tabular}}
\end{table*}

\paragraph{Deviations.}
\begin{itemize}
\item \emph{Serving the fine-tuned model.} vLLM could not load the LoRA adapter for this model's architecture, so, as the preregistration allowed, we served the base model with the adapter merged into its weights (the same adapter, not retrained; token counts checked against the server before each run).
\item \emph{Scoring order.} Local scoring was the bottleneck, so from the terse-prompting experiment on we first scored only the patches the primary tests need (final patch, first pass, twice passed, last pass) and scored the remaining intermediate patches later. Primary tests are unaffected; anytime curves use the complete scores.
\item \emph{Asking the agent.} The first round of queries capped answers at 8 tokens. The model answered in its agent format (``THOUGHT: The tests pass\ldots'') and was cut off before stating a decision; only 3 of 439 answers could be parsed. We kept these results, raised the cap to 256 tokens, parsed the last occurrence of SUBMIT or CONTINUE, and re-asked all 439 queries with the question unchanged; the thresholds were unchanged.
\item \emph{Excluded run.} In the combined online experiment one run's recorded snapshot did not match its submitted patch and was excluded by a rule fixed in advance.
\item \emph{Environments.} Sphinx tests run the command that \texttt{tox.ini} specifies directly instead of through tox (\cref{app:envs}).
\item \emph{Second scaffold.} Besides the sandbox changes in \cref{app:envs}, a socket path that was too long for long task names made 23 runs on the new repositories fail before the agent started. We fixed it and re-ran them; the failed attempts are kept. SWE-agent counts its call limit one request later than mini-SWE-agent.
\item \emph{Gemma-4.} Its smoke test failed our format threshold in mini-SWE-agent's text format (23\% of turns without a command). In tool-call format it passed that threshold, but our per-request token counts did not match the server's, because the model emits a thinking segment that the server re-renders. Our preregistration then required stopping. We preregistered a second attempt in tool-call format that takes the token counts from the prompt tokens the server returns with each response; it passed its smoke test (all requests matched exactly, against 20 of 60 with the tokenizer endpoint) and is the one we report.
\end{itemize}

\section{Stopping Details}
\label{app:stop}
\paragraph{Replay and scoring.} Each recorded command is re-executed in the executable sandbox at the task's base commit with the same 60-second limit, and the tracked-file \texttt{git diff} is recorded after every command. Distinct non-empty patches are scored like SWE-bench: the model patch is applied, test files touched by the test patch are restored, the test patch is applied, and the task's test command runs; the SWE-bench 4.1.0 log parser decides FAIL\_TO\_PASS and PASS\_TO\_PASS, with Django's two-line docstring format joined into one line. Cost up to step $k$ sums the three streams of requests $1..k$.

\paragraph{Self-test rule.} A command counts as a test run if it matches \texttt{runtests.py}, \texttt{bin/test}, \texttt{pytest} or \texttt{unittest}. Its output counts as passing if it contains \texttt{OK}, ``$N$ passed'' or ``tests finished: $N$ passed'' and no failure marker (\texttt{FAILED}, \texttt{FAIL:}, \texttt{ERROR:}, a traceback, ``failed'' other than ``0 failed'', or an error count). The rule fires at the first such command while the working tree has a non-empty patch; the online version forces a submission of that patch before the next model request.

\paragraph{After the rule fires (historical $D$).} Of the 414 runs in which the rule fired, 278 had a passing patch both at that point and at the end, 122 had a failing patch at both, 11 went from passing to failing and 3 from failing to passing. Stopping removed 39\% of the cost of these runs.

\paragraph{Loss decomposition and a risk-controlled rule (historical $D$).}\label{app:loss} %
Let $L$ and $G$ indicate that stopping destroys or rescues a success; they split $-r\cdot u\cdot\delta$ in \cref{eq:decomp}, and the change in resolution is $E[G]-E[L]$. For the rules that stop at the $k$-th pass on the same patch ($k=4,3,2,1$), we chose $k$ on the 460 calibration runs by a fixed-sequence Learn-then-Test procedure (preregistered): test $E[L]\le0.02$ from the most conservative rule on, with a one-sided 95\% task-clustered bootstrap bound, and keep the last rule that passes. The bound for $k=3$ was 0.028, so the procedure chose $k=4$, which saved only 4\% on the 78 new-repository tasks and the 178 further tasks and failed the preregistered 15\% saving. On those held-out sets, $k=2$ kept the measured $E[L]$ at 0.013 and 0.006 (bounds 0.038 and 0.011) for savings of 21\% and 19\%. In SWE-agent the same procedure, calibrated on the 115 fresh tasks, also chose $k=4$; on the 78 new-repository tasks it saved 32\% with $E[L]=0.013$, but the bound (0.026) exceeded the budget, so the preregistered confirmation failed there too, while $k=2$ lost 0.032 (bound 0.058). Risk control with a 2\% budget is thus feasible but conservative at these sample sizes; it matters more in SWE-agent, where more stopped runs would have been fixed later.

\paragraph{Learned policy.} Features: a constant, $\log(1+k)$, a non-empty patch, $\log(1+{}$commands since the last patch change$)$, the number of patch changes (capped at 5, scaled), whether a test ran since the last change, whether it passed, their product with a non-empty patch, whether evidence was given, and the kind of the last command (read, search, edit, script). Two L2-regularized logistic regressions predict passing now and at the end, fitted by IRLS with five-fold cross-fitting by task. The expected remaining cost is the mean over training runs in the same ten-step bucket. The policy stops when $\lambda\,p_{\text{now}} \ge \lambda\,p_{\text{end}} - \text{cost}_{\text{remaining}}$, swept over $\lambda$ from $5\times10^3$ to $10^6$ tokens.

\section{Cost Accounting}
\label{app:cost}
We price runs as in \cref{eq:cost}. Server-reported cache hits can include prefixes that a concurrent run with an identical prompt prefix has just computed; a deployed agent has no such sibling. We therefore cap each request's hits at the prompt length of the run's own previous request. In runs where we recorded token ids, we instead use the longest common token prefix with the previous request, measured with the serving model's tokenizer.

\section{Environments for Seven Further Repositories}
\label{app:envs}
For pytest, Sphinx, xarray, pylint, requests, seaborn and Flask we built one Python environment per repository version in SWE-bench Verified (43 in total), following the SWE-bench 4.1.0 specifications: the Python version, pinned packages, requirement files at the version's environment-setup commit, and the specification's \texttt{sed} edits. We install the project once with its extras to collect its dependencies, then remove the project but keep its metadata and console scripts. A path file points the interpreter at the task checkout. Four fixes were needed. First, we add setuptools, which our virtual environments lacked. Second, pytest checkouts lack the version file that installation generates, so a finder supplies it when the file is absent. Third, xarray needs the optional dependencies that its official tests exercise. Fourth, Sphinx needs a docutils version of its era. Sphinx tests run the command in \texttt{tox.ini} directly, because tox's own interpreter link escapes our virtual environment. The gold patch passes the hidden tests on 78 of 106 tasks; most failures are requests tests that need network access.

\paragraph{Sandbox details for the two scaffolds.}
The repository in the sandbox belongs to the host user, so git refuses commands from the sandbox user. In the mini-SWE-agent runs we left this as it was: most runs hit it once or twice, in 3--5\% of their commands, and we always took patches on the host. SWE-agent needs git to set up the repository and to submit, so in its sandbox we marked the repository as safe. SWE-agent's tools run on a separate Python 3.12 with their dependencies preinstalled, because the sandbox has no network and some task environments use Python 3.8. Its submit command stages all files, so we take every snapshot as the difference of the tracked files against the base commit, which equals the unstaged diff that mini-SWE-agent's snapshots use.

\section{Terse Prompting}
\label{app:terse}
In an earlier pilot, adding ``Keep the THOUGHT section to at most two short sentences.'' to the system prompt saved 13--23\% of cost. We preregistered a test on the 115 fresh tasks, two runs each. Cost fell by only 3.9\%, output tokens by 3.7\%, and the number of steps barely changed (49.8 against 50.6). Resolution changed by $+0.4$ points. Neither criterion was met. These runs serve as a second sample for the stopping rules in \cref{sec:results}.

\section{Comparison with a Turn Cap}
\label{app:turncap}
This comparison retains the historical $D$ convention. \citet{gao2026more} cap an agent's turns at the 75th percentile of its unrestricted run lengths and terminate runs that reach the cap. Anytime evaluation prices this rule exactly on our five sets of Qwen3.6-27B runs. Taken as specified, it saves 13--21\% and loses 8--16 points, with every interval below zero. Even if the run instead submits its current patch at the cap, it loses up to 4.3 points for the same saving. On the same runs, stopping at the first pass saves 26--31\% and the twice-passed rule 18--23\%, each within 3 points of running to the end under $D$. This bound does not hold for the corrected fresh-task first-pass result, which is $-3.9$ points under $D'$. A cap ignores whether the patch already works. Their dynamic variant, which grants extra turns with a message to the agent, changes the rest of the run and is therefore not a prefix rule; it cannot be priced from recorded runs.

\section{Combining Stopping with Context Management}
\label{app:combine}
Stopping also interacts with context management. Break-even masking removes old observations when their cached price over the expected remaining requests $R$ exceeds the cost of recomputing the rest of the prompt (companion work). We ran it together with the twice-passed rule online on the 193 tasks, two runs each, and applied last-pass selection afterwards (preregistered). At $c=1$ the combination saved 33.2\%, about seven points more than stopping and selection alone. At $c=0.1$ it saved only 3.5\% and failed our criterion. Masking raised uncached input from 49K to 79K tokens per run. Its estimate of the remaining steps assumed runs that go to the end, so the recomputation was never paid back once the rule shortened them. Stopping and selection alone saved 21.8\% at $c=0.1$ with no change in resolution. Replayed on existing masked runs (preregistered), stopping took most of what masking saves, and at $c=0.1$ masking made stopped runs more expensive. When caching is cheap, stop and select without masking; when it is not, add masking.

\section{Telling, Asking and Judging the Agent}
\label{app:ask}
\paragraph{Telling or asking the agent instead.}
Could the agent stop itself? We replaced the last step of the official workflow (test edge cases) with an instruction to submit as soon as the fix works and the relevant tests pass, and ran it twice on the 115 fresh tasks (preregistered). Cost fell by only 8.4\%, short of the $-10\%$ we required, and runs shortened by about one step. The external rule still saved 36\% on top of it. The extra work is therefore not mainly prompted; it is how the model verifies. Nor does the agent know when it is done. At its first passing self-test we appended one short query to the cached conversation, asking whether it would submit now (preregistered; the query costs little at $c=0.1$). It answered SUBMIT in only 6--7\% of 439 cases, and mostly ignored the question and issued its next verification command. Yet in 61--72\% of the cases where it chose to continue, the patch it had already passed the hidden tests. Stopping on its answer therefore changed cost by less than 1\%.

\paragraph{An external judge does no better.}
We also asked a judge in a fresh context, the same model given only the issue, the current patch and the passing test's command and output, whether to submit; we stopped at its first yes (preregistered, the 193 tasks, two runs each, with the judge's calls priced as uncached input). It stopped 37\% of the runs and saved 14.0\% for $-0.5$ points, so the quality criterion failed. The twice-passed rule, on the same runs, saved 20.8\% for the same change in resolution. The judge was cautious rather than lenient: it asked to continue on 160 of the 272 patches that already passed the hidden tests. In SWE-agent it saved 32.5\% for $-2.3$ points, against 54.2\% for $-2.8$ with the rule. Neither the agent nor a fresh judge of the same model beats a rule that just counts passing tests.

\section{Other Models}
\label{app:models}
\paragraph{Which agents it applies to.}
The rule needs an agent that runs the repository's tests. Qwen3-Coder-30B-A3B, run twice on the same 115 fresh tasks (preregistered), issued only 34 recognizable test commands in 230 runs and verified its fixes with its own scripts. The rule never fired and saved nothing. Gemma-4-26B-A4B (preregistered, same tasks), one of the models outside the Qwen family, behaved the same way: it ran a recognizable test command in 33 of 230 runs and saw one pass in 6, so the rules fired in at most 2\% of runs and changed neither cost nor resolution. GLM-4.7-Flash (preregistered, same tasks) ran tests in 73 of 230 runs but saw them pass in only 14, and half of its runs hit the step limit; the twice-passed rule fired in 3\% of runs and saved 2.4\%. Qwen3.6-27B rarely ran a recognizable test command on SymPy tasks either (1 of 48 fresh runs fired). Qwen3.8-27B, a newer model of the same architecture (preregistered, same 115 tasks), does run the tests, but its runs are already short (32 steps against 51). There the twice-passed rule fired in 30\% of runs and saved 13.6\% for $-0.9$ points, narrowly missing both criteria. Last pass never applied, because this model did not change its patch after its tests passed. Qwen3.5-9B (preregistered, same tasks) runs long (88 steps) and tests often, but its tests rarely pass. The twice-passed rule fired in 11\% of runs and saved only 3.1\% without loss ($+0.4$ points). What these rules save depends on how long a model keeps working after it has a verified patch (\cref{sec:decomp}). For Qwen3.5-9B the staged-patch correction raises final resolution from 0.465 to 0.522; the earlier claim about a late decline must therefore be interpreted under its original $D$ convention.

\paragraph{A non-Qwen model that runs tests.}
DeepSeek-V4-Flash provides a different pattern on the same fresh tasks (preregistered; $D'$). Twice passed saves 13.5\% for $+0.9$ resolution points. The primary test fails its cost criterion, despite meeting the quality criterion. First-pass stopping, a preregistered secondary comparison, saves 19.0\% for $+1.7$ points, meeting both numerical thresholds. This supports a cost--quality trade-off outside Qwen, not a universal twice-passed rule (\cref{app:deepseek}).

\paragraph{Scripts as checks.} To reach agents that verify with their own scripts, we also counted a direct run of a Python script as a check, passing when it exits with 0 and prints no traceback, error or failure (preregistered, the same recorded runs, no new runs). Submitting at the first passing check fired in 43\% of Gemma-4-26B-A4B's runs, saved 12.2\% and gained 2.2 points, because many of its runs otherwise end at the step or context limit; submitting once a patch passed two checks saved 4.8\%. Neither met the 10\% cost criterion. For Qwen3-Coder both fired in at most 7\% of runs. For Qwen3.6-27B the same two rules lost 8.7 and 5.7 points: a passing script is a weaker signal than the repository's tests.

\section{Second Scaffold: SWE-agent}
\label{app:scaffold}
\paragraph{SWE-agent works longer after a pass, and some of it pays.}
In SWE-agent, Qwen3.6-27B ran 93 and 85 steps on the fresh tasks and new repositories, against 51 and 52 in mini-SWE-agent, and ran the repository's tests 12--16 times per run instead of 4--9. It resolved more (0.809 and 0.692 against 0.752 and 0.673 under $D\prime$) at 2.6 times the cost. After its first pass it ran 51 and 44 more commands, two thirds of the cost, and changed the patch again in nearly half of the runs (final patch unchanged in 55\% and 58\%, against 80\% in mini-SWE-agent). That work was not idle: from the first pass to the end, 24 failing patches became passing and 3 passing ones failed. The twice-passed rule fired in 88\% of the runs and saved 54.2\% of the cost, but lost 2.8 points of resolution, so the preregistered quality criterion failed. Stopping at the first pass saved 59.9\% for $-5.4$ points. Last-pass selection in SWE-agent adds only $0.3$ points; its snapshots already use a base-relative diff and are unaffected by this correction. Requiring four passes on the same patch, the rule that risk control selected here (\cref{app:loss}), saved 35\% and 32\% for $0.0$ and $-0.6$ points; there, 7--9\% of the runs that passed a self-test did so on a failing patch they later fixed, against 2--6\% in mini-SWE-agent. The rules save money in both scaffolds; whether the work they cut was useful depends on how the scaffold spends the steps after a pass. Combined online with break-even masking from companion work, stopping and selection saved 33\% at $c=1$ but only 3.5\% at $c=0.1$, because masking assumed runs that go to the end (\cref{app:combine}).

\section{Fine-Tuning in SWE-agent}
\label{app:sweft}
\paragraph{In SWE-agent it learns the rule's trade-off.}
We repeated the fine-tuning in SWE-agent on its runs of the 115 fresh tasks (preregistered): the same labels, with the context and tool definitions SWE-agent actually sent, a second example that confirms the submission when SWE-agent asks for a review, and a 48K-token limit for its longer prompts. On the 78 tasks of the new repositories, two runs each, the fine-tuned agent cut cost by 56.9\% and changed resolution by $-3.2$ points, so the quality criterion failed. It matched the twice-passed rule on the base model's runs almost exactly ($+0.1\%$ cost, $-0.6$ points), and confirmed at the review prompt in 145 of 156 runs. Other labels did not remove the loss (preregistered): stopping at the first truly passing self-test, by the hidden tests of the training tasks, gave $-51\%$ and $-4.5$ points; stopping at the fourth pass (0.6 points lost offline) gave $-48\%$ and $-2.6$ points: the model stopped earlier than the rule.

\section{Anytime Curves: Two Further Observations}
\label{app:curves}
Retrieved evidence in the prompt moves the anytime curve earlier in steps (0.51 versus 0.41 resolved at step 20) but not higher at the end (0.646 versus 0.652). We do not pursue restarts: when a task's first run failed, a second independent run succeeded only 16\% of the time. Cost varies between runs of the same task, as \citet{bai2026ai} observed across models, but the extra spending does not buy resolution. The costliest of four runs of a task costs a median 1.6 times the cheapest, and when two runs of a task end differently, the costlier one is the resolved one in only 36--58\% of pairs across our three task sets.

\section{What the Agent Tests When the Rule Fires}
\label{app:tests}
We classified the first passing self-test of 893 Qwen3.6-27B runs on all four sets by its test target, using the hidden FAIL\_TO\_PASS tests only for this analysis. In 52\% of the runs the agent ran the module or file that holds the hidden tests, and the patch passed them 72\% of the time. In 21\% it ran other tests of the repository (60\%), and in 5\% only a script of its own (60\%). The remaining 22\%, mostly SymPy, could not be classified (70\%). Running the right tests thus makes a pass more trustworthy, but not reliable: even then, 28\% of the patches fail. Most false passes, however, cost nothing to stop on: in mini-SWE-agent 25--34\% of the runs that passed a self-test did so on a failing patch, but only 2--6\% later reached a passing one (\cref{app:loss}). Even tests that fail before a fix and pass after it can admit incorrect patches \citep{tan2026beyond}. A stricter online rule that counts only passes of tests related to the edited files, by their names, fired less and saved less without raising resolution.

\section{Exact-Advantage Labels and Training Seeds}
\label{app:rl}
\begin{figure}[h]
\centering
\figorbox{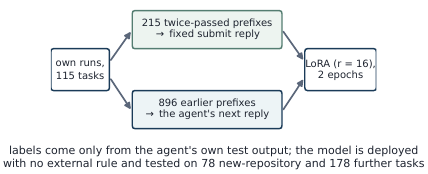}{\linewidth}
\caption{Stopping supervision for Qwen3.6-27B. Prefixes of the agent's own runs, cut where its tests had passed twice on the same patch or earlier, with the targets shown; prefixes longer than 32K tokens are dropped.}
\label{fig:stopsft}
\end{figure}

\paragraph{Labels.} On the same 460 training runs as the twice-passed labels, every passing self-test that is not the last command is a decision point $k$. Its advantage is $A_k=\lambda(Q_k-Q_T)+\mathrm{cost}_T-\mathrm{cost}_k$ at $c=0.1$, $p=4$, with $\lambda=10^6$ in the same units. A decision point with $A_k>0$ becomes a stop example; otherwise the agent's actual next reply is the target. Points after the first stop are unreachable and are not used. In the \emph{hidden-test} version $Q$ is the verdict of the hidden tests on the patch, used only for labels of training tasks. In the \emph{own-test} version $Q$ is whether the agent's last self-test on that patch passed, which uses the agent's later tests but no hidden tests. The same background continue examples as for twice passed fill the set to 1{,}149 examples, the size of the twice-passed set before 38 examples longer than 32K tokens were dropped. The submit reply was changed to a neutral one (``the tests I ran pass''). $\lambda$ exceeded the saving at every decision point, so a label says: stop at a pass unless later work repairs the patch (hidden tests) or the agent's later tests overturn the pass (own tests). Both versions stop in 273 of the 460 runs, 253--257 of them at the first pass, against 215 stops at the twice-passed point.

\paragraph{Results.} On the 178 tasks, two runs each, the own-test version resolved 0.626 at a cost of 121K, the hidden-test version 0.629 at 114K, against 0.632 and 175K for the base model and 0.657 and 127K for the twice-passed agent. About half of their runs submitted without a passing self-test (172 and 170 of 356). On the 78 new-repository tasks they saved 23.7\% and 30.0\% and changed resolution by $+3.8$ and $-3.2$ points. Pooled over 256 tasks, the hidden-test version saved 8.1\% more than the twice-passed agent and resolved 3.7 points less; the own-test version was within 1.3\% and 1.8 points of it.

\paragraph{Training seeds ($D\prime$ reanalysis).} We retrained the twice-passed agent with five further seeds (LoRA initialization, dropout and data order; data, recipe and sampling seeds unchanged) and ran each on the 178 tasks and the 78 new-repository tasks, two runs per task (\cref{tab:seeds}). To separate training from evaluation noise, we also ran the first model and the base model again with new sampling seeds, so both have four runs per task. Against the base model's four runs, every seed saves 28--33\% over 256 tasks; quality changes range from $-3.1$ to $+1.2$ points under $D'$. The equal-weight mean is $-1.2$ points at a 30.0\% saving. Uncertainty was assessed with crossed seed--task resampling and, separately, task-only resampling. The quality margin is not established. On the 178-task split the mean is $-0.3$ points; on 78 new-repository tasks it is $-3.3$. The original checkpoint's matched second evaluation over all 256 tasks changes quality by $-3.7$ points. These results retain the original resampling and thresholds; the historical variance diagnostics below were computed under $D$.

\begin{table}[h]
\centering
\caption{Stop distillation with six training seeds against four runs per task of the base model: change in local resolution (points) and cost at $c=0.1$. Seed 1 has four runs per task, seeds 2--6 two. Below: each evaluation of seed 1 against the base model's runs of the same period, and the base model's second evaluation against its first.}
\label{tab:seeds}
\small
\setlength{\tabcolsep}{4pt}
\begin{tabular}{lrrrr}
\toprule
 & \multicolumn{2}{c}{178 tasks} & \multicolumn{2}{c}{256 tasks} \\
Training seed & resolved & cost & resolved & cost \\
\midrule
1 & $-0.4$ & $-30.7\%$ & $-1.1$ & $-31.4\%$ \\
2 & $-0.7$ & $-29.4\%$ & $-3.1$ & $-30.0\%$ \\
3 & $+2.4$ & $-29.0\%$ & $+1.2$ & $-28.1\%$ \\
4 & $-2.4$ & $-34.5\%$ & $-2.3$ & $-33.2\%$ \\
5 & $-1.3$ & $-30.6\%$ & $-2.1$ & $-29.3\%$ \\
6 & $+0.4$ & $-27.6\%$ & $+0.2$ & $-28.0\%$ \\
\midrule
Mean, 1--6 & $-0.3$ & $-30.3\%$ & $-1.2$ & $-30.0\%$ \\
\midrule
1, runs 0--1 & $+2.5$ & $-27.6\%$ & $+1.6$ & $-27.7\%$ \\
1, runs 2--3 & $-3.4$ & $-33.5\%$ & $-3.7$ & $-34.9\%$ \\
Base, runs 2--3 & $+2.0$ & $+5.6\%$ & $+1.2$ & $+7.5\%$ \\
\bottomrule
\end{tabular}
\end{table}

\section{Precision of Offline Evaluation}
\label{app:power}
Under $D'$, on the 178 tasks of \cref{sec:learned}, each a mean over two runs, the standard error of a change in resolution is 0.40 points for the twice-passed rule against running to the end on the same runs, 0.63 for the first-pass rule, 1.61 for the distilled agent against the base agent (independent runs) and 1.51 between two evaluations of the base agent. Comparing policies on the same recorded runs is thus worth 6.5--16.5 times as many tasks. \Cref{tab:power} converts these variances into the tasks needed to detect a change with 80\% power at a true change of zero.
\begin{table}[h]
\centering
\caption{Tasks needed to detect a change in resolution ($80\%$ power, two-sided $5\%$), from the per-task variances on the 178 tasks.}
\label{tab:power}
\small
\resizebox{\columnwidth}{!}{%
\begin{tabular}{lrrrr}
\toprule
Comparison & runs per task & 1 point & 2 points & 3 points \\
\midrule
Twice passed vs.\ end, same runs & 2 & 221 & 56 & 25 \\
First pass vs.\ end, same runs & 2 & 554 & 139 & 62 \\
Distilled vs.\ base agent, online & 2 & 4{,}735 & 1{,}184 & 527 \\
Distilled vs.\ base agent, online & 8 & 1{,}184 & 296 & 132 \\
\bottomrule
\end{tabular}}
\end{table}

\section{Decomposing the Distilled Agents}
\label{app:decomp}
\begin{table}[t]
\caption{Historical $D$ diagnostic (not recomputed under $D'$): where a distilled agent's change in resolution comes from (points against the base agent run to the end; held-out tasks, two runs each: 178 for mini-SWE-agent against the base agent's runs 0--1, 78 for SWE-agent). Teacher: the labelling rule on the base agent's runs. Stopping part: the fine-tuned model deciding only at the passes of the same runs. Online: the fine-tuned agent's own runs. $e$: share of runs where the stopping part submits a different patch than the teacher. Imitation error is stopping part minus teacher; the rest is online minus stopping part, which includes sampling noise ($\approx$1.6 points, \cref{app:power}). XL: mean of six training seeds; $^{*}$exploratory. This table is not a decomposition of the corrected six-seed result.}
\label{tab:diag}
\centering
\footnotesize
\setlength{\tabcolsep}{2pt}
\resizebox{\linewidth}{!}{%
\begin{tabular}{@{}llrrrr@{}}
\toprule
Label & Setting & Teacher & Stopping & Online & $e$ \\
\midrule
Twice passed (XL) & mini, Qwen3.6 & $-0.6$ & $-0.1$ & $+1.0$ & 0.02 \\
Twice passed & mini, Qwen3.8 & $-0.3$ & $-0.6$ & $-0.3$ & 0.02 \\
Exact advantage, hidden / own tests & mini, Qwen3.6 & $+0.6$ / $-0.3$ & $-0.3$ / $-0.3$ & $-0.3$ / $-0.6$ & 0.01 / 0.03 \\
Random / matched position & mini, Qwen3.6 & -- & $-0.3$ / $-0.3$ & $-61.0$ / $-12.1$ & 0.01 / 0.03 \\
Twice passed & SWE-agent & $-3.2$ & $-3.9$ & $-2.6$ & 0.04 \\
Hidden-test labels & SWE-agent & $+0.6$ & $-3.9$ & $-3.2$ & 0.09 \\
Fourth pass & SWE-agent & $-1.9$ & $-3.9$ & $-1.9$ & 0.22 \\
Fourth pass + continue examples & SWE-agent & $-1.9$ & $-1.3$ & $-1.9^{*}$ & 0.13 \\
\bottomrule
\end{tabular}}
\end{table}

The stopping part is measured by giving each fine-tuned model, at every passing self-test of the base agent's recorded runs, the context it would have seen, at the online sampling settings (temperature 0.7, top-$p$ 0.8, a fixed seed per decision); it stops where its reply submits. In mini-SWE-agent we queried 609 decisions per model on the 178 tasks (381 for Qwen3.8-27B), in SWE-agent 1{,}051 on the 78 new-repository tasks; no query failed. For the six training seeds the imitation error was $+0.3$ to $+0.6$ points and the online residual $-1.1$ to $+3.7$. In SWE-agent the imitation errors were $-0.6$ for twice passed, $-4.5$ for hidden-test labels, $-1.9$ for fourth-pass labels and $+0.6$ with continue examples; the residuals were within $\pm2$ points with substantial statistical uncertainty. The stopping part saves 24--25\% in mini-SWE-agent against 19\% for its teacher, and the online agents 26--33\%: they also submit more often without any passing test. With continue examples the fourth-pass student's stopping part lost 1.4 points per run at both temperatures (0.7 points at temperature 0), the rule's own loss, but its $e$ of 0.127 missed the preregistered threshold of 0.10.

\section{Which Pass to Wait For}
\label{app:hazard}
Let $Q_j$ and $C_j$ be the resolution and mean cost when every run stops at its $j$-th pass on one patch. Waiting for one more pass pays when a resolved task is worth more than $\lambda_j=(C_{j+1}-C_j)/(Q_{j+1}-Q_j)$ tokens at $c=0.1$, $p=4$. In mini-SWE-agent, under $D'$, waiting rarely pays: on the 178 replication tasks (four runs each) every further pass changed resolution by at most 0.1 points and the first pass is optimal for every $\lambda$ from $10^5$ to $10^7$ tokens; on the 78 new-repository tasks the first pass is optimal up to $\lambda=10^6$. On the 115 fresh tasks pooled over four runs per task (460 runs, rather than the 230-run fresh-task comparison in \cref{tab:stop}), the first pass loses 3.0 points and the estimated optimum moves from the first pass ($\lambda\le3\times10^5$) to the second, the fourth and, at $\lambda=10^7$, running to the end. For DeepSeek-V4-Flash the first pass has the highest estimated utility at every tested $\lambda$. These are descriptive optima on the recorded runs, not held-out policy guarantees. In SWE-agent each pass up to the fourth repaired 0.6--2.6 points, with $\lambda_j$ between 0.8 and 8.4 million tokens; the optimal pass moves from the first ($\lambda\le3\times10^5$) to the fourth or fifth ($\lambda=10^7$). For Qwen3.8-27B it moves from the first to the fourth. The rule to distill is thus set by the scaffold's repair rate after a pass and by what a resolved task is worth.

\section{Patch Definition and Scoring Sensitivity}
\label{app:metric}
\paragraph{Three conventions.}
$D$ is the recorded \texttt{git diff}, comparing the worktree to the index. It excludes staged edits and newly created files. $D'$ is the preregistered limited correction: only when $D$ is empty and the recorded \texttt{diff\_all}, restricted to files tracked at the base commit, is non-empty do we replace the end patch. For those runs we replay commands and use base-relative tracked-file snapshots, accepting them only when the reconstructed final state matches. Non-empty but partially staged $D$ remains unchanged. Consequently $D'$ is not a complete base-relative patch protocol. The correction is applied equally across arms and does not alter model inputs or thresholds.

P2 instead uses the parsed submitted patch when it is non-empty and differs from $D$, falling back to $D$ when no submitted patch is available. It is a sensitivity analysis, not an exact implementation of the official mini-SWE-agent protocol: new files remain excluded and the fallback is part of our definition. The fine-tuned agent seldom writes a patch file, so this fallback is used asymmetrically across arms. P1, which scores only the submitted patch, is recorded in the audit but is not our primary metric.

\paragraph{Submitted-patch sensitivity.}
Under P2, the base agent's resolved rate falls by 3.4 points on derivation runs, 6.1 on fresh tasks and 3.7 on the 178-task distillation evaluation, whereas the fine-tuned arm on the latter set is unchanged. This is not universal: on new-repository runs the base rate rises by 2.6 points. Many submitted patches fail to apply; the audit found that roughly half of these application failures can be rescued by the official harness's fuzzy fallback. Neither patch convention changes the token cost of an executed run. Their asymmetry makes P2 unsuitable for interpreting the fine-tuned model's gain as a pure stopping effect; all P2 comparisons are kept in the audit.

\paragraph{Staged changes and replay.}
The original correction scope contains 107 distinct runs. All 107 CPU replays reproduce the restricted final state; no run is silently removed after a failed first attempt. Final resolved rates rise from 0.649 to 0.657 on derivation runs, 0.739 to 0.752 on fresh tasks and 0.641 to 0.673 on new repositories. The Gemma-4-31B rate rises from 0.358 to 0.428 when averaging runs (0.426 when averaging tasks first, because one task has only one included run). Its twice-passed rule still never triggers and saves nothing. Final-run token costs do not change. Rule costs can change if a corrected prefix changes the trigger; for example, the derivation first-pass saving changes from 23.5\% to 23.9\%.

\begin{table}[h]
\centering\small
\caption{Quality changes (points) under the original and corrected conventions. Test verdicts use the original prespecified thresholds. All corrected scores are local.}
\begin{tabular}{@{}lrr@{}}
\toprule
Comparison & $D$ & $D'$ \\
\midrule
Fresh first pass & $-3.0$ & $-3.9$ \\
New-repository last pass & $+3.2$ & $0.0$ \\
Replication twice passed & $0.0$ & $-1.0$ \\
Replication last pass & $+1.3$ & $+0.3$ \\
Six-seed distillation, 256 tasks & $-0.3$ & $-1.2$ \\
\bottomrule
\end{tabular}
\end{table}

\paragraph{Changed conclusions.}
For the 386-run twice-passed replication, the quality estimate changes from $0.0$ to $-1.0$ points, and the corrected test no longer establishes non-inferiority. The 22.8\% cost saving still meets its criterion. The corrected tests no longer establish a positive gain from last-pass selection on either new repositories or the replication, so those earlier claims are withdrawn. The six-seed quality estimate changes from $-0.3$ to $-1.2$ points; non-inferiority remains unestablished. The paired reanalysis preserves both scoring conventions.

\paragraph{Official scoring and historical diagnostics.}
Earlier Windows official evaluations received $D$, sharing its staged-edit omission with local scoring. Their agreement cannot validate $D'$. Official reports have returned for all 307 distinct corrected final and intermediate patches in that package. Its gold self-check passes on 92 of 93 tasks; the remaining task fails. These reports cover the exported subset rather than all corrected experiments, including the later DeepSeek correction. Coverage is detailed in \cref{app:officialreturn}. The primary stopping table, six-seed table and updated anytime figures use $D'$; diagnostic probes, older ablations and the historical preregistration table retain $D$ unless explicitly marked as corrected. They cannot be combined with corrected effects as an exact decomposition. This audit narrows the evidence for loss-free stopping and motivates treating patch identity as part of the evaluation protocol.

\section{DeepSeek Cross-Family Evaluation}
\label{app:deepseek}
The primary rule was twice passed, with a cost upper confidence bound of at most $-10\%$ and a quality lower bound of at least $-2$ points. The rules and thresholds were frozen before inference. Of 230 completed runs, one failed final-state consistency, leaving 229 runs on 115 tasks. We average repeats within tasks and use the same 2,000 task bootstrap resamples (seed 20261001). All scores here are local, not official harness scores.

The audit found 11 additional runs requiring the empty-$D$ staged-edit correction; all 11 replays matched their restricted final state. Scoring infrastructure failures were retained separately and retried, never counted as model failures. Corrected final resolution is 0.752, versus 0.717 under $D$ and 0.683 under P2. The difference from run-averaged rates reflects the one excluded repeat.

\begin{table}[h]
\centering\footnotesize
\setlength{\tabcolsep}{1.5pt}
\caption{DeepSeek quality changes in points. Each convention uses its own run-to-end baseline.}
\begin{tabular}{@{}lrrr@{}}
\toprule
Rule & $D$ & $D'$ & P2 \\
\midrule
First pass & $+3.0$ & $+1.7$ & $+3.5$ \\

Twice passed & $+1.7$ & $+0.9$ & $+3.0$ \\

Last pass & $+3.9$ & $+2.6$ & $+0.9$ \\

\bottomrule
\end{tabular}
\end{table}

Across conventions, first-pass cost changes by $-19.0\%$, twice-passed cost by $-13.5\%$, and last-pass cost is unchanged. The primary twice-passed cost test is unsupported under all three conventions. Under $D'$, first-pass stopping meets both numerical thresholds as a secondary result; its positive quality point estimate is not evidence of a strictly positive gain. Last-pass selection has a positive local gain under $D'$, unlike the corrected Qwen replication. It remains a secondary, model-specific result and does not restore the withdrawn Qwen claim. P2 is a hybrid scoring convention, not the official submission protocol (\cref{app:metric}).

\section{Coverage of Additional Official Evaluations}
\label{app:officialreturn}
We received four additional Windows SWE-bench harness result packages. The archive checksums match the supplied manifest. The table reports coverage of distinct non-empty patches, separately from the gold-patch self-check on each task. Missing reports are unknown outcomes, not failures. Empty model patches remain unresolved under the existing protocol.
\begin{center}
\setlength{\tabcolsep}{2pt}
\begin{tabular}{lrrr}
\toprule
Package & Reports & Gold pass & Tasks \\
\midrule
Seeds/repeats & 1,531/1,532 & 255 & 256 \\
Base repeats & 433/433 & 255 & 256 \\
Corrected patches & 307/307 & 92 & 93 \\
Additional runs & 1,338/1,341 & 430 & 451 \\
\bottomrule
\end{tabular}
\end{center}
The first three packages each contain one task whose gold self-check fails. In the fourth, seven gold self-checks fail and fourteen are missing. Its manifest records a patch-application failure, a test timeout and an image-download failure for the three missing model-patch reports. These records remain missing rather than being inferred from the status messages. The paired official analyses below use matched tasks with valid gold checks and complete reports. Local estimates retain their original denominators and labels. The correction package covers 107 original runs and their intermediate patches, not the later DeepSeek corrections.

\paragraph{Agreement on identical patches.}
We compared local and official results using the task ID and patch hash, excluding tasks without a passing official gold self-check. In the staged-edit correction package, all 305 eligible distinct patches agree with local scoring. In the training/sampling package, 1,522 of 1,523 comparable patches agree; in the base-agent repeat package all 430 comparable patches agree, with one further patch lacking a local score. In the additional context/stopping package, 554 of 564 comparable patches agree. Another 691 eligible patches lack an unambiguous local score and are not included in this agreement calculation. Thus the returned reports validate the exported correction subset, but do not establish agreement for every experiment or patch convention.

\paragraph{Paired official comparisons.}
We retain only tasks with passing gold checks and all required reports in both conditions, average repeats within tasks, and use 2,000 paired task bootstrap samples (seed 20261001). These are supplementary official analyses on their stated task subsets. They do not replace local results on a different denominator. Missing reports remain unknown, and the later DeepSeek corrections are outside this package.
\begin{center}
\small
\setlength{\tabcolsep}{2pt}
\begin{tabular}{lrr}
\toprule
Comparison & Tasks & Change (pp) \\
\midrule
XL vs. XR (r2--r3) & 255 & $-3.9$ \\
Matched-stop vs. XR & 178 & $-12.1$ \\
\bottomrule
\end{tabular}
\end{center}

For the six-seed distillation comparison, we require all four base-agent repeats, all four repeats of the original distilled checkpoint, and both repeats of each remaining training seed. This leaves 254 tasks (177 Django/SymPy and 77 new-repository tasks). One task is excluded for a failed gold check and another for a missing seed report. With training seeds weighted equally and tasks resampled within repository-set strata, crossed seed/task bootstrap gives a resolution difference of $-1.1$ points and a cost change of $-30.1\%$, using the same measured token costs at $(c,p)=(0.1,4)$. The cost criterion is met, but the $-2$-point quality margin is not. This agrees with the qualitative conclusion of the local reanalysis.